\documentclass[11pt,a4paper]{article}
\usepackage[hyperref]{emnlp2020}
\usepackage{times}
\usepackage{latexsym}

\usepackage{microtype}

\aclfinalcopy 
\def\aclpaperid{2962} 

\usepackage{subcaption}
\usepackage{tabularx}
\usepackage{booktabs}

\usepackage{graphicx}
\usepackage{enumitem}
\usepackage{csvsimple}
\usepackage{dsfont}
\usepackage{pifont}
\usepackage{amsthm}
\usepackage[ruled,vlined]{algorithm2e}
\usepackage{xspace}
\usepackage{amsmath}
\usepackage{arydshln}
\usepackage{multirow}
\usepackage{colortbl}
\usepackage{float}
\usepackage{multicol}
\usepackage{array}
\usepackage{amssymb}
\usepackage{soul}

\setlist{nolistsep}

\newcommand{\ours}{\textsc{SciFact}\xspace}

\newcommand{\sysname}{\textsc{VeriSci}\xspace}

\newcommand{\fever}{\textsc{Fever}\xspace}
\newcommand{\eraser}{\textsc{ERASER}\xspace}
\newcommand{\snopes}{UKP Snopes\xspace}

\newcommand{\supports}{\textsc{Supports}\xspace}
\newcommand{\refutes}{\textsc{Refutes}\xspace}
\newcommand{\notenough}{\textsc{NoInfo}\xspace}

\newcommand{\support}{\textsc{Support}\xspace}
\newcommand{\refute}{\textsc{Refute}\xspace}

\newcommand{\numclaims}{1,409\xspace}
\newcommand{\numpapers}{5,183\xspace}

\newcommand{\labelkappa}{0.75\xspace} 
\newcommand{\rationalekappaagree}{0.71\xspace} 
\newcommand{\nagreement}{232\xspace}              

\newcommand{\bert}{\textsc{BERT}\xspace}
\newcommand{\scibert}{\textsc{SciBERT}\xspace}

\newcommand{\bioroberta}{BioMedRoBERTa\xspace }
\newcommand{\robertabase}{RoBERTa-base\xspace}
\newcommand{\robertalarge}{RoBERTa-large\xspace}

\newcommand{\cls}{\textsc{CLS}\xspace}
\newcommand{\sep}{\textsc{SEP}\xspace}

\newcommand{\yhat}{\widehat{y}}

\newcommand{\Shat}{\widehat{S}}
\newcommand{\shat}{\widehat{s}}

\newcommand{\Ehat}{\widehat{\cE}}

\definecolor{NiceGreen}{RGB}{125,204,110}
\definecolor{NiceRed}{RGB}{222,0,0}
\definecolor{c1}{rgb}{0.12, 0.46, 0.70}
\definecolor{c2}{rgb}{1.0, 0.50, 0.05}
\definecolor{c3}{rgb}{0.17,0.63,0.17}

\newcommand{\abstractLabelOnly}{$\textrm{Abstract}_{\textrm{Label-Only}}$\xspace}
\newcommand{\abstractLabelRationale}{$\textrm{Abstract}_{\textrm{Label+Rationale}}$\xspace}
\newcommand{\sentenceSelectionOnly}{$\textrm{Sentence}_{\textrm{Selection-Only}}$\xspace}
\newcommand{\sentenceSelectionLabel}{$\textrm{Sentence}_{\textrm{Selection+Label}}$\xspace}

\newcommand{\componentone}{\textsc{AbstractRetrieval}\xspace}
\newcommand{\componenttwo}{\textsc{RationaleSelection}\xspace}
\newcommand{\componentthree}{\textsc{LabelPrediction}\xspace}

\newcommand{\thinmid}{\arrayrulecolor{black!30}\midrule}
\newcommand{\thickmid}{\arrayrulecolor{black}\midrule}

\newcolumntype{L}[1]{>{\raggedright\let\newline\\\arraybackslash\hspace{0pt}}m{#1}}
\newcolumntype{C}[1]{>{\centering\let\newline\\\arraybackslash\hspace{0pt}}m{#1}}
\newcolumntype{R}[1]{>{\raggedleft\let\newline\\\arraybackslash\hspace{0pt}}m{#1}}

\definecolor{c1}{RGB}{21,156,0}
\newcommand{\supportsCovid}[1]{{\color{c1} {#1}}}

\definecolor{c2}{RGB}{218,0,0}
\newcommand{\refutesCovid}[1]{{\color{c2} {#1}}}

\definecolor{c3}{RGB}{127,127,127}

\definecolor{lightblue}{RGB}{212, 235, 255}
\definecolor{salmon}{RGB}{255, 164, 168}
\definecolor{lightgreen}{RGB}{177, 231, 171}
\definecolor{lightyellow}{RGB}{255, 255, 148}
\sethlcolor{lightblue}

\newcommand\hlc[2]{\sethlcolor{#1} \hl{#2}}

\newcommand{\cA}{{\mathcal{A}}}

\newcommand{\cE}{{\mathcal{E}}}

\newcommand{\cR}{{\mathcal{R}}}

\title{Fact or Fiction: Verifying Scientific Claims}

\author{David Wadden$^\mathbf{\dagger}$\thanks{\xspace \xspace Work performed during internship with the Allen Institute for Artificial Intelligence.} \quad
Shanchuan Lin$^\mathbf{\dagger}$ \quad
Kyle Lo$^\mathbf{\ddagger}$ \quad
Lucy Lu Wang$^\mathbf{\ddagger}$ \\
{\bf Madeleine van Zuylen}$^\mathbf{\ddagger}$ \quad
{\bf Arman Cohan}$^\mathbf{\ddagger}$ \quad
{\bf Hannaneh Hajishirzi}$^\mathbf{\dagger\ddagger}$\\
  $^\mathbf{\dagger}$
  University of Washington, Seattle, WA, USA \\
  $^\mathbf{\ddagger}$
  Allen Institute for Artificial Intelligence, Seattle, WA, USA \\
  {\tt\small \{dwadden,linsh,hannaneh\}@cs.washington.edu} \\
  {\tt\small \{kylel,lucyw,madeleinev,armanc\}@allenai.org}
}

\date{}

\begin{document}
\maketitle

\begin{abstract}

We introduce scientific claim verification, a new task to select abstracts from the research literature containing evidence that \supports or \refutes a given scientific claim, and to identify rationales justifying each decision. To study this task, we construct \ours, a dataset of 1.4K expert-written scientific claims paired with evidence-containing abstracts annotated with labels and rationales. We develop baseline models for \ours, and demonstrate that simple domain adaptation techniques substantially improve performance compared to models trained on Wikipedia or political news. We show that our system is able to verify claims related to COVID-19 by identifying evidence from the CORD-19 corpus. Our experiments indicate that \ours will provide a challenging testbed for the development of new systems designed to retrieve and reason over corpora containing specialized domain knowledge. Data and code for this new task are publicly available at \url{https://github.com/allenai/scifact}. A leaderboard and COVID-19 fact-checking demo are available at \url{https://scifact.apps.allenai.org}.

\end{abstract}
\section{Introduction} \label{sec:introduction}

\begin{figure}[t]
  \centering

  \includegraphics[width=\columnwidth]{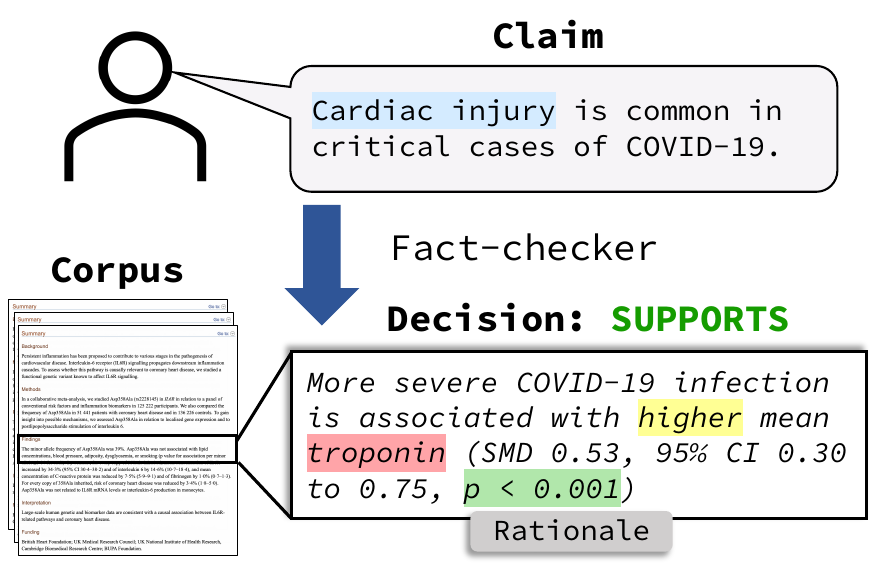}
  \caption{A scientific claim, supported by evidence identified by our system. To correctly verify this claim, the system must possess background knowledge that \emph{\hlc{salmon}{troponin}} is a protein found in cardiac muscle and that elevated levels of \emph{\hlc{salmon}{troponin}} are a marker of \emph{\hlc{lightblue}{cardiac injury}}. In addition, it must be able to reason about directional relationships between scientific processes: replacing \emph{\hlc{lightyellow}{higher}} 
  with \emph{\hlc{lightyellow}{lower}}
  would cause the rationale to \refute the claim rather than \support it. Finally, the system should 
  interpret \emph{\hlc{lightgreen}{$p < 0.001$}} as an indication that the reported finding is statistically significant.}

  \label{fig:teaser}
\end{figure}

Due to rapid growth in the scientific literature, it is difficult for researchers -- and the general public even more so -- to stay up to date on the latest findings. This challenge is especially acute during public health crises like the current COVID-19 pandemic, due to the extremely fast rate at which new findings are reported and the risks associated with making decisions based on outdated or incomplete information. As a result, there is a need for automated tools to assist researchers and the public in evaluating the veracity of scientific claims.

\begin{table*}[t]
  \setlength{\tabcolsep}{.25em}
  \footnotesize
  \centering

  \begin{tabularx}{\textwidth}{ X }
    \toprule
    \textbf{Claim 1}: Lopinavir / ritonavir have exhibited favorable clinical responses when used as a treatment for coronavirus. \\
    \midrule
    \textbf{Supports}: \dots \supportsCovid{\emph{Interestingly, after lopinavir/ritonavir (Kaletra, AbbVie) was administered, $\beta$-coronavirus viral loads significantly decreased and no or little coronavirus titers were observed.}} \\
    \midrule
    \textbf{Refutes}: \refutesCovid{\emph{The focused drug repurposing of known approved drugs (such as lopinavir/ritonavir) has been reported failed for curing SARS-CoV-2 infected patients.}} It is urgent to generate new chemical entities against this virus \dots \\
    \bottomrule
    \vspace{1mm} \\
    \toprule
    \textbf{Claim 2}: The coronavirus cannot thrive in warmer climates. \\
    \midrule
    \textbf{Supports}: \supportsCovid{\emph{...most outbreaks display a pattern of clustering in relatively cool and dry areas...This is because the environment can mediate human-to-human transmission of SARS-CoV-2, and unsuitable climates can cause the virus to destabilize quickly...}} \\
    \midrule
    \textbf{Refutes}: \refutesCovid{\emph{...significant cases in the coming months are likely to occur in more humid (warmer) climates, irrespective of the climate-dependence of transmission and that summer temperatures will not substrantially limit pandemic growth}}. \\
    \bottomrule
  \end{tabularx}

  \caption{Evidence identified by our system as supporting and refuting two claims concerning COVID-19.}
  \label{tbl:covid_example}

\end{table*}

\emph{Fact-checking} -- a task in which the veracity of an input \emph{claim} is verified against a corpus of documents that \emph{support} or \emph{refute} the claim -- has been studied to combat the proliferation of misinformation in political news, social media, and on the web~\cite{Thorne2018FEVERAL,Hanselowski2019ARA}. However, verifying scientific claims poses new challenges to both dataset construction and effective modeling. While political claims are readily available on fact-checking websites and can be verified by crowd workers, annotators with extensive domain knowledge are required to generate and verify scientific claims.

In addition, NLP systems for scientific claim verification must possess additional capabilities beyond those required to verify factoid claims. For instance, to verify the claim shown in Figure \ref{fig:teaser}, a system must have the ability to access scientific background knowledge, reason over increases and decreases in quantities or measurements, and make sense of specialized statistical language.

In this paper, we introduce the task of scientific claim verification to evaluate the veracity of scientific claims against a scientific corpus. Table \ref{tbl:covid_example} presents some examples. To facilitate research on this task, we construct \ours, an expert-annotated dataset of \numclaims scientific claims accompanied by abstracts that support or refute each claim, and annotated with rationales \cite{Lei2016RationalizingNP} justifying each \supports\ / \refutes decision. To create the dataset, we develop a novel annotation protocol in which annotators re-formulate naturally occurring claims in the scientific literature -- \emph{citation sentences} -- into atomic scientific claims. Using citation sentences as a source of claims both speeds the claim generation process and guarantees that the topics discussed in \ours are representative of the research literature. In addition, citation links indicate the exact documents likely to contain evidence necessary to verify a given claim.

We establish performance baselines on \ours with an approach similar to \citet{DeYoung2019ERASERAB}, which achieves strong performance on the \fever claim verification dataset~\cite{Thorne2018FEVERAL}. Our baseline is a pipeline system which retrieves abstracts related to an input claim, uses a \bert-based \cite{Devlin2019BERTPO} sentence selector to identify rationale sentences, and labels each abstract as \supports, \refutes, or \notenough with respect to the claim. We demonstrate that our baseline can benefit from training on claims from domains including Wikipedia articles and politics.

We showcase the ability of our model to verify expert-written claims concerning the novel coronavirus COVID-19 against the newly-released CORD-19 corpus \cite{Wang2020CORD19TC}. Expert annotators judge retrieved evidence to be plausible for 23 of 36 claims.\footnote{We emphasize that our model is a \emph{research prototype} and should not be used to make any medical decisions whatsoever.} Our results and analyses demonstrate the importance of the new task and dataset to support significant future research in this domain.

In summary, our contributions include:
(1) We introduce and formalize the scientific claim verification task.
(2) We develop a novel annotation protocol to generate and verify 1.4K naturally-occurring claims about scientific findings.
(3) We establish strong baselines on this task, and identify substantial opportunities for improvement at all stages of the modeling pipeline.
(4) We demonstrate the efficacy of our system in a real-world case study verifying claims about COVID-19 against the research literature.

\section{Background and task definition}
As illustrated in Figure \ref{fig:teaser}, scientific claim verification  is the task of identifying evidence from the research literature that \supports or \refutes a given scientific claim. Table \ref{tbl:covid_example} shows the results of our system  applied to claims about the novel coronavirus COVID-19. For each claim, the system identifies relevant scientific abstracts, and labels the relation of each abstract to the claim as either \supports or \refutes. Verifying scientific claims is challenging and requires domain-specific background knowledge -- for instance, in order to identify the evidence supporting Claim 1 in  Table~\ref{tbl:covid_example}, the system must determine that a reduction in coronavirus viral load indicates a favorable clinical response, even though this fact is never mentioned.

\vspace{.1cm}
\noindent{\bf Scientific claims} \label{sec:claims} In \ours, a scientific claim is an \emph{atomic verifiable statement} expressing a finding about one aspect of a scientific entity or process, which can be verified from a single source.\footnote{Requiring annotators to search multiple sources increases cognitive burden and decreases annotation quality.}
For instance,  ``\emph{The $R_0$ of the novel coronavirus is 2.5}'' is valid, but opinion-based statements like ``\emph{The government should require people to stand six feet apart to stop coronavirus}'' are not. Compound claims like ``\emph{Aerosolized coronavirus droplets can travel at least 6 feet and can remain in the air for 3 hours}'' should be split into two atomic claims.

Claims in \ours  are \emph{natural} -- they are derived from citation sentences, or \emph{citances} \cite{Nakov2004CitancesC}, that occur naturally in scientific articles. This is similar to political fact-checking datasets such as \snopes \cite{Hanselowski2019ARA}, which use political fact-checking websites as a source of natural claims. On the other hand, claims in the popular \fever dataset \cite{Thorne2018FEVERAL} are \emph{synthetic}, since they are created by annotators by mutating sentences from the Wikipedia articles that will serve as evidence.

\vspace{.1cm}
\noindent{\bf Supporting and refuting evidence} \label{sec:evidence}
In most fact-checking work, claims are assigned a global truth label based on the entirety of the available evidence. For example in \fever, the claim ``\emph{Barack Obama was the $44^{th}$ President of the United States}'' can be verified using Wikipedia as an evidence source.

While \ours claims are indeed verifiable assertions about scientific findings, accurately assigning a global truth label to a scientific claim (given a fixed scientific corpus) requires a systematic review by a team of experts. In this work we focus on the simpler task of assigning \supports or \refutes relations to individual \emph{claim-abstract pairs}.

Each \supports or \refutes relation between claim and abstract must be justified by at least one \emph{rationale}. A rationale is a minimal collection of sentences which, taken together as premises in the context of the abstract, can reasonably be judged by a domain expert as implying the claim. Rationales facilitate the development of \emph{interpretable models} which not only have the ability to make label predictions, but can also identify the exact sentences that are necessary for their decisions.

\section{The \ours dataset} \label{sec:dataset}

\label{sec:dataset_overview}

The \ours dataset consists of \numclaims scientific claims\footnote{\ours is comparable in size to recent scientific datasets for tasks such as QA (e.g. PubMedQA~\cite{Jin2019PubMedQAAD} has 1,000 questions), and information extraction (e.g. SciERC~\cite{Luan2018MultiTaskIO} has 500 annotated abstracts).} verified against a corpus of \numpapers abstracts. Abstracts that support or refute each claim are annotated with rationales.  We describe our corpus creation and annotation process.


\subsection{Data source and corpus construction}
\label{sec:data_source}

\begin{figure}[t]
  \centering

  \includegraphics[width=\columnwidth, height=6.5cm]{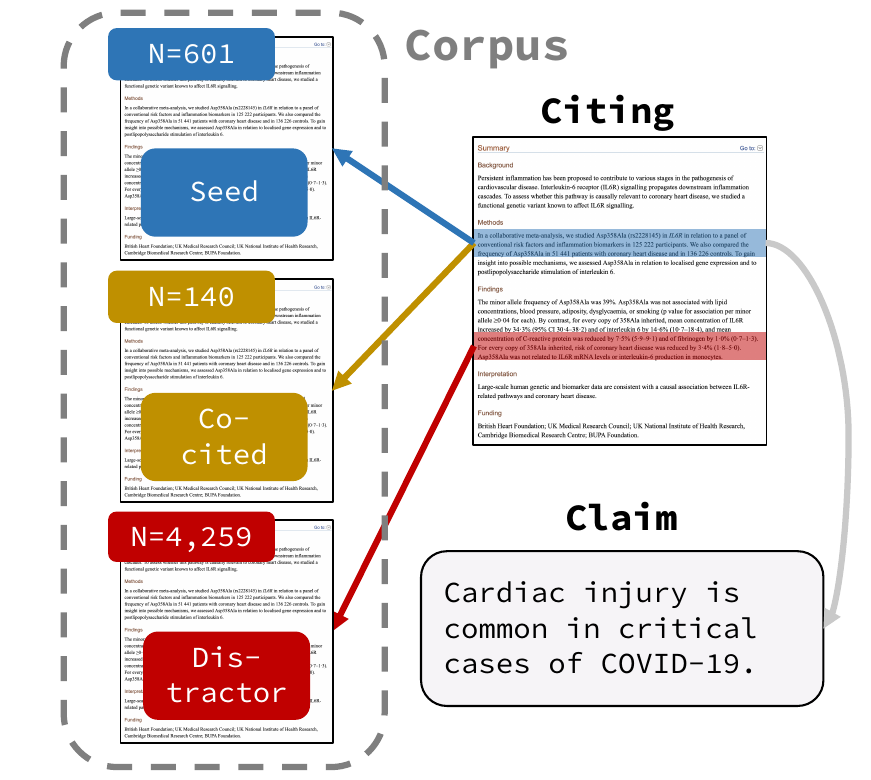}
  \caption{Corpus construction. Citing abstracts are identified for each seed document. A claim is written based on the source citance in the citing abstract.}
  \label{fig:corpus}

\end{figure}

\begin{figure}[t]
  \centering
  \includegraphics[width=\columnwidth]{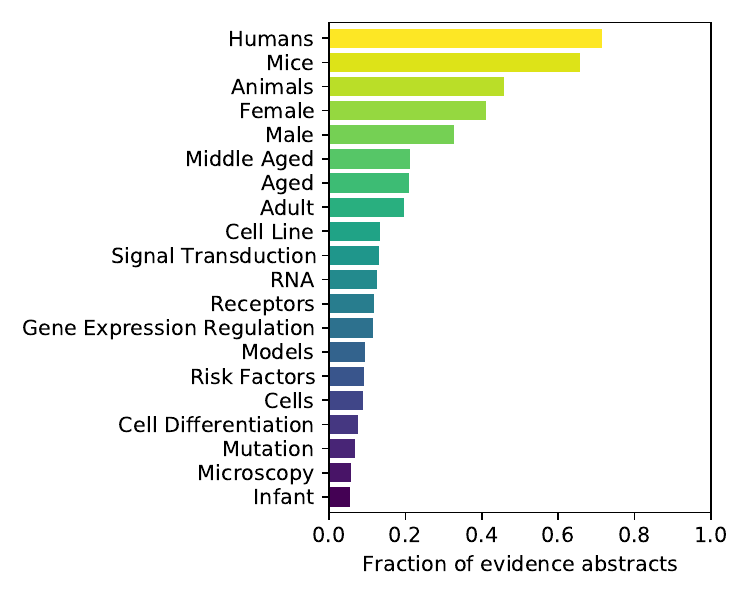}

  \caption{Most frequently occurring Medical Subject Headings (MeSH) terms (y-axis) among cited abstracts. MeSH is a controlled vocabulary used for indexing articles in PubMed. Topics range from clinical trial reports (``Humans'', ``Risk Factors'') to molecular biology (``Cell Line'', ``RNA'').}
  \label{fig:mesh_terms}
\end{figure}

To construct \ours, we use S2ORC \cite{Lo2019GORCAL}, a publicly-available corpus of millions of scientific articles. To ensure that documents in our dataset are of high quality, we randomly sample articles from a manually curated collection of well-regarded journals spanning domains from basic science (e.g., \emph{Cell}, \emph{Nature}) to clinical medicine (e.g., \emph{JAMA}, \emph{BMJ}). The full list of journals is included in Appendix \ref{sec:data_source_appendix}. We restrict to articles with at least 10 citations. The resulting collection is referred to as our \textit{seed} set. We use the S2ORC citation graph to sample \emph{source citances} from \emph{citing articles} which cite these seed articles. If a citance cites other articles not in the seed set, we refer to these as \emph{co-cited} articles and add them to the corpus, as depicted in Figure \ref{fig:corpus}. The content of the cited abstracts encompasses a diverse array of topics within biomedicine, as shown in Figure \ref{fig:mesh_terms}. The majority of citances used for \ours cite only the seed article (no co-cited articles), as we found in initial annotation experiments that these citances tended to yield specific, easy-to-verify claims.

To expand the corpus, we identify five papers cited in the same paper as each source citance but in a different paragraph, and add these to the corpus as \emph{distractor abstracts}. These abstracts often discuss similar topics to the evidence documents, increasing the difficulty of abstract retrieval and making our metrics more accurately reflect the system's performance on a large research corpus.


\subsection{Claim writing} \label{sec:claim_writing}
\noindent {\bf Annotation}
Annotators are shown a source citance in the context of an article, and are asked to write up to three claims based on the content of the citance; see Appendix \ref{sec:annotation_examples} for an example. This results in \emph{natural} claims because the annotator does not see the cited article's abstract -- the \emph{cited abstract} -- at the time of claim writing. Annotators are asked to skip citances that do not make statements about specific scientific findings.

The claim writers included four experts with background in scientific NLP, fifteen undergraduates studying the life sciences, and four graduate students (doctoral or medical) in the life sciences. Detailed information on the annotator training process can be found in Appendix \ref{sec:annotator_training}. The claim-writing interface is shown in Appendix \ref{sec:annotation_interfaces}.

\vspace{.1cm} \noindent{\bf Claim negation} Unless the authors of the source citance were mistaken, cited articles should provide supporting evidence for the claims made in a citance. To obtain examples where an abstract \refutes a claim, an NLP expert wrote negations of existing claims, taking precautions not to bias the negations by using obvious keywords like ``not'' \cite{Schuster2019TowardsDF, Gururangan2018AnnotationAI}. In  \S\ref{sec:pipeline_components}, we demonstrate that a ``claim-only'' verification model performs poorly, suggesting that the negation process did not introduce severe artifacts.


\subsection{Claim verification}
\label{sec:claim_verification}

\noindent {\bf Annotation} For each claim, all of the claim's cited abstracts are annotated for evidence. Annotators are shown a single claim - cited abstract pair, and asked to label the pair as \supports, \refutes, or \notenough. Although our task definition allows for a single claim to be both supported and refuted (by different abstracts) -- an occurrence we observe on real-world COVID-19 claims (\S\ref{sec:covid}) -- this never occurs in our dataset. Each claim has a single label. Counts for each label are shown in Table \ref{tbl:label_counts}. Overall, the annotators found evidence in 63\% of cited abstracts.  If the annotator assigns a \supports or \refutes label, they must also identify all rationales as defined in \S\ref{sec:evidence}. Table \ref{tbl:evidence_counts} provides statistics on the number of sentences per rationale, the number of rationales per claim / abstract pair, and the number of evidence abstracts per claim.  No abstract has more than 3 rationales for a given claim, and all rationales consist of at most three sentences. Rationales in \ours are mutually exclusive. 28 rationales contain non-contiguous sentences.

The verifiers included three NLP experts, five life science undergraduates, and five graduate students studying life sciences. Annotators verified claims that they did not write themselves. Annotation guidelines are provided in Appendix \ref{sec:annotation_interfaces}.

\ours claims are verified against abstracts rather than full articles since (1) abstracts can be annotated more scalably, (2) evidence is found in the abstract in more than 60\% of cases, and (3) previous attempts at full-document annotation suffered from low annotator agreement (\S\ref{sec:related_scinlp}).

\begin{table}[t]
  \setlength{\tabcolsep}{.25em}
  \footnotesize
  \centering

  \begin{subtable}{\linewidth}
    \begin{tabular}{L{0.12\linewidth} C{0.19\linewidth} C{0.19\linewidth} C{0.19\linewidth} C{0.15\linewidth}}
      \toprule
      Fold &  \supports &  \notenough &  \refutes &   All \\
      \midrule
      Train &       332 &              304 &      173 &   809 \\
      Dev   &       124 &              112 &       64 &   300 \\
      Test  &       100 &              100 &      100 &   300 \\
      \midrule
      All   &       556 &              516 &      337 &  1409 \\
      \bottomrule
    \end{tabular}
    \subcaption{Distribution of claim labels in \ours.}
    \label{tbl:label_counts}
  \end{subtable}

  \vspace{0.25em}

  \begin{subtable}{\linewidth}
    \begin{tabular}{L{0.48\linewidth} C{0.1\linewidth} C{0.1\linewidth} C{0.1\linewidth} C{0.1\linewidth}}
      \toprule
      {} &    0 &     1 &    2 &   3+ \\
      \midrule
      Cited abstracts per claim     &    - &  1278 &   86 &   45 \\
      Evidence abstracts per claim &  516 &   830 &   37 &   26 \\
      Rationales per abstract         &    - &   552 &  290 &  153 \\
      Sentences per rationale       &    - &  1542 &   92 &   11 \\
      \bottomrule
    \end{tabular}
    \subcaption{Evidence counts at various levels of granularity. For example, Column 2 of the row ``Rationales / abstract'' indicates that 290 claim / abstract pairs are supported by 2 distinct rationales.}
    \label{tbl:evidence_counts}
  \end{subtable}

  \caption{Statistics on claim labels, and the number of evidence abstracts and rationales per claim.}
  \label{tbl:evidence_stats}

\end{table}

\vspace{.1cm} \noindent{\bf Quality}
We assign \nagreement claim-abstract pairs for independent re-annotation. The label agreement is \labelkappa Cohen's $\kappa$, comparable with the 0.68 Fleiss' $\kappa$ reported in \citet{Thorne2018FEVERAL}, and 0.70 Cohen's $\kappa$ reported in \citet{Hanselowski2019ARA}. To measure rationale agreement, we treat each sentence as either classified as ``part of a rationale'' or ``not part of a rationale'' and compute sentence-level agreement. The resulting Cohen's $\kappa$ is \rationalekappaagree.

\section{The \ours task} \label{sec:task_definition}
\noindent {\bf Task Formulation} The inputs to our task are a scientific claim $c$ and a corpus of abstracts $\cA$.
All abstracts $a \in \cA$ are labeled as $y(c, a)\in\{$\supports, \refutes, \notenough$\}$ with respect to a claim $c$.  The abstracts that either \support or \refute $c$ are referred to as \emph{evidence abstracts} for $c$, denoted as $\cE(c)$. 
Each evidence abstract $a \in \cE(c)$ is annotated with rationales. A single rationale $R_i$ is a collection of sentences $\{r_1(c, a), \dots, r_{m}(c, a)\}$, where $m$ is the number of sentences in rationale $R_i$.
We denote the set of all rationales as $\cR(c, a) = \{R_1(c, a), \dots, R_n(c, a)\}$.

Given a claim $c$ and a corpus $\cA$, the system must predict a set of evidence abstracts $\Ehat(c)$. For each abstract $a \in \Ehat(c)$, it must predict a label $\yhat(c, a)$, and a collection of \emph{rationale sentences} $\Shat(c, a) = \{\shat_1(c, a), \dots, \shat_{\ell}(c, a)\}$. Note that although the gold annotations may contain multiple separate rationales, to simplify the prediction task we only require the model to predict a single collection of \emph{rationale sentences}; these sentences may encompass multiple gold rationales.

\vspace{.1cm} \noindent {\bf Task Evaluation} \label{sec:evaluation}
We evaluate the task at two levels of granularity. For \emph{abstract-level} evaluation, we assess the model's ability to identify the abstracts that support or refute the claim. For \emph{sentence-level} evaluation, we evaluate the model's performance at identifying the sentences sufficient to justify the abstract-level predictions. We conduct evaluations in both the ``Open'' \fever-style \cite{Thorne2018FEVERAL} setting where the evidence abstracts must be retrieved, and the ``Oracle abstract'' \eraser-style \cite{DeYoung2019ERASERAB} setting where the gold evidence abstracts $\cE(c)$ are provided.

\vspace{.1cm} \noindent {\bf Abstract-level evaluation} is inspired by the \fever score. Given a claim $c$, a predicted evidence abstract $a \in \Ehat(c)$ is \emph{correctly labeled} if (1) $a$ is a gold evidence abstract for $c$, and (2) The predicted label is correct: $\yhat(c, a) = y(c, a)$. It is \emph{correctly rationalized} if, in addition, the predicted rationale sentences contain a gold rationale, i.e., there exists some gold rationale $R_i(c, a) \subseteq \Shat(c, a)$.

Like \fever, which limits the maximum number of predicted rationale sentences to five, \ours limits to three predicted rationale sentences. Overall performance is measured by the micro-F1 of the precision and recall over the correctly-labeled and correctly-rationalized evidence abstracts. We refer to these evaluations as \abstractLabelOnly and \abstractLabelRationale, respectively.

\vspace{.1cm} \noindent {\bf Sentence-level evaluation} measures performance in identifying individual rationale sentences. Unlike the abstract-level metrics, this evaluation penalizes the prediction of extra rationale sentences.

A predicted rationale sentence $\shat(c, a)$ is correctly \emph{selected} if (1) It is a member of some gold rationale $R_i(c, a)$, (2) all other sentences from the same gold rationale $R_i(c, a)$ are among the predicted $\Shat(c, a)$, and (3) $\yhat(c, a) \neq \notenough$\footnote{Condition (3) eliminates  rationale sentences which were identified by the rationale selector, but proved insufficient to justify a final \supports\ / \refutes decision}. It is correctly \emph{labeled} if, in addition, the abstract $a$ is correctly labeled: $\yhat(c, a) = y(c, a)$.

Overall performance is measured by the micro-F1 of the precision and recall of correctly-selected and correctly-labeled rationale sentences, denoted \sentenceSelectionOnly and \sentenceSelectionLabel. For sentence-level evaluation, we do not limit the number of predicted rationale sentences, since the evaluation penalizes models that over-predict.

\section{\sysname: Baseline model} \label{sec:baselines}
We develop a baseline (referred to as \sysname) that takes a claim $c$ and corpus $\cA$ as input, identifies evidence abstracts $\Ehat(c)$, and predicts a label $\yhat(c,a)$ and rationale sentences $\Shat(c, a)$ for each $a \in \Ehat(c)$.  Following the ``\bert-to-\bert''  model presented in \citet{DeYoung2019ERASERAB,Soleimani2019BERTFE}, \sysname\ is a pipeline of three components:

\begin{enumerate}[leftmargin=*]
    \item \componentone retrieves $k$ abstracts with highest TF-IDF similarity to the claim.
    \item \componenttwo identifies rationale sentences $\Shat(c, a)$ for each abstract.
    \item \componentthree makes the final label prediction $\yhat(c, a)$.
\end{enumerate}


 \noindent {\bf Rationale selection} \label{sec:rationale_selection}
Given a claim $c$ and abstract $a$, we train a model to predict $z_i \triangleq \mathds{1}[a_i \textrm{ is a rationale sentence}]$ for each sentence $a_i$ in $a$. For each sentence, we encode the concatenated sequence $w_i = [a_i, \sep, c]$ using a \bert-style language model and predict a score $\tilde{z}_i = \sigma[f(\cls(w_i))]$, where $\sigma$ is the sigmoid function, $f$ is a linear layer and $\cls(w_i)$ is the \cls token from the encoding of $w_i$. 
We train the model on pairs of claims and their cited abstracts and minimize cross-entropy loss between $z_i$ and $\tilde{z}_i$. For each claim, we use cited abstracts labeled \notenough, as well as non-rationale sentences from abstracts labeled \supports and \refutes as negative examples. To make predictions, we select all sentences $a_i$ with $\tilde{z_i} > t$ as rationale sentences, where $t \in [0, 1]$ is tuned on the dev set (Appendix \ref{sec:parameters}).



\vspace{.1cm} \noindent {\bf Label prediction} \label{sec:entailment}
Sentences identified by the rationale selector are passed to a separate \bert-based model to make the final
labeling decision. Given a claim $c$ and abstract $a$, we concatenate the claim and the predicted rationale sentences $u = [\shat_1(c, a), \dots \shat_{\ell}(c, a), \sep, c]$\footnote{We truncate the rationale input if it exceeds the \bert token limit. $c$ is never truncated.}, and predict $\tilde{y}(c, a) = \phi[f(\cls(u))]$, where $\phi$ is the softmax function, and $f$ is a linear layer with three outputs representing the \{\supports, \refutes, \notenough\} labels. We minimize the cross-entropy loss between $\tilde{y}(c, a)$ and the true label $y(c, a)$.

We train the model on pairs of claims and their cited abstracts using gold rationales as input. For cited abstracts labeled \notenough, we choose the $k$ sentences from the cited abstract with highest TF-IDF similarity to the claim as input rationales. For prediction, we use the predicted rationale sentences $\Shat(c, a)$ as input and predict $\hat{y}(c, a) = \textrm{argmax }\tilde{y}(c, a)$. \notenough is predicted for abstracts with no rationale sentences.

We experimented with a label prediction model which encodes entire abstracts via the Longformer \cite{Beltagy2020LongformerTL}, and makes predictions using the document-level \cls token. Performance was not competitive with our pipeline setup, likely because the label predictor struggles to identify relevant information when given full abstracts.
\section{Experiments}

In our experiments, we (1) analyze the performance of each individual component of \sysname, (2) evaluate full task performance in both the ``Oracle abstract'' and ``Open'' settings, (3) present promising results verifying claims about COVID-19 using \sysname, and (4) discuss some modeling challenges presented by the dataset.





\subsection{Pipeline components} \label{sec:pipeline_components}
We examine the effects of different training datasets, sentence encoders, and model inputs on the performance of the \componenttwo and \componentthree modules. The \componenttwo module is evaluated on its ability to select rationale sentences given gold abstracts\footnote{Our \fever-trained \componenttwo module achieves 79.9 sentence-level F1 on the \fever test set, virtually identical to 79.6 reported in \citet{DeYoung2019ERASERAB}.}. The \componentthree module is evaluated on its 3-way label classification accuracy given gold rationales from cited abstracts. Cited abstracts labeled \notenough are included in the evaluation. These abstracts have no gold rationale sentences; as in \S\ref{sec:baselines}, we provide the $k$ most similar sentences from the abstract as input (more details in Appendix \ref{sec:model_details}).

\begin{table}[t]
  \footnotesize
  \setlength{\tabcolsep}{0.5em}
  \centering

  \begin{tabularx}{\columnwidth}{l *{3}{c}  *{3}{c}}
    \toprule
 & \multicolumn{3}{c}{\textsc{Rational-Select.}} & \multicolumn{3}{c}{\textsc{Label-Pred.}} \\

    \midrule
    \textbf{Training data} & P & R & F1 & \multicolumn{3}{c}{\textsc{Acc.}} \\

    \arrayrulecolor{black!30}\midrule

    \fever         &      41.5 &   57.9 &  48.4 &  \multicolumn{3}{c}{67.6} \\
    \snopes        &      42.5 &   62.3 &  50.5 &  \multicolumn{3}{c}{71.3} \\
    \ours          &      73.7 &   70.5 &  \textbf{72.1} &  \multicolumn{3}{c}{75.7} \\
    \fever + \ours &      72.4 &   67.2 &  69.7 &  \multicolumn{3}{c}{\textbf{81.9}} \\

    \arrayrulecolor{black}
    \midrule

    \textbf{Sentence encoder} & P & R & F1 & \multicolumn{3}{c}{\textsc{Acc.}} \\
    \arrayrulecolor{black!30}\midrule

    \scibert      &      74.5 &   74.3 &  \textbf{74.4} &  \multicolumn{3}{c}{69.2} \\
    \bioroberta   &      75.3 &   69.9 &  72.5 &  \multicolumn{3}{c}{71.7} \\
    \robertabase  &      76.1 &   66.1 &  70.8 &  \multicolumn{3}{c}{62.9} \\
    \robertalarge &      73.7 &   70.5 &  72.1 &  \multicolumn{3}{c}{\textbf{75.7}} \\

    \arrayrulecolor{black}
    \midrule

    \textbf{Model inputs} & P & R & F1 & \multicolumn{3}{c}{\textsc{Acc.}} \\
    \arrayrulecolor{black!30}\midrule
    Claim-only         &      - &   - &  - &  \multicolumn{3}{c}{44.5} \\
    Abstract-only         &   60.1 &   60.9 &  60.5 &  \multicolumn{3}{c}{53.3} \\

    \arrayrulecolor{black}
    \bottomrule
  \end{tabularx}

  \caption{Comparison of different training datasets, encoders, and model inputs for \componenttwo and \componentthree, evaluated on the \ours dev set. The claim-only model cannot select rationales.}
  \label{tbl:components}

\end{table}

\vspace{.1cm} \noindent{\bf Training Data}
We train on (1) \fever, (2) \snopes, (3) \ours, and (4) \fever pretraining followed by \ours fine-tuning. \robertalarge \cite{Liu2019RoBERTaAR} is used as the sentence encoder.

\vspace{.1cm} \noindent{\bf Sentence encoder}
We fine-tune \scibert \cite{Beltagy2019SciBERTPC}, \bioroberta \cite{Gururangan2020DONTST}, \robertabase, and \robertalarge. \ours is used as training data.

\begin{table*}[t]
  \footnotesize
  \setlength{\tabcolsep}{0.45em}
  \centering

  \begin{tabularx}{1.02\linewidth}{*{3}{l} *{6}{c} *{1}{c} *{6}{c}}


    \toprule

    & & & \multicolumn{6}{c}{\textbf{Sentence-level}} & & \multicolumn{6}{c}{\textbf{Abstract-level}} \\

     & & & \multicolumn{3}{c}{\textbf{Selection-Only}} & \multicolumn{3}{c}{\textbf{Selection+Label}} & & \multicolumn{3}{c}{\textbf{Label-Only}} & \multicolumn{3}{c}{\textbf{Label+Rationale}}  \\

    \cmidrule(lr){4-6} \cmidrule(lr){7-9} \cmidrule(lr){11-13} \cmidrule(lr){14-16}

    Retrieval & Model &  & P & R & F1 & P & R & F1 & & P & R & F1 & P & R & F1 \\

    \midrule

    \multirow{4}{*}{\shortstack[l]{\textbf{Oracle}\\ \textbf{abstract}}} &

    Oracle rationale & 1 & 100.0 & 80.5 & $89.2_{2.1}$ & 89.6 & 72.2 & $79.9_{3.0}$ &  & 90.1 & 77.5 & $83.3_{2.4}$ & 90.1 & 77.5 & $83.3_{2.4}$ \\

    \arrayrulecolor{black!30}\cmidrule(lr){2-16}

 & Zero-shot & 2 & 42.5 & 45.1 & $43.8_{2.0}$ & 36.1 & 38.4 & $37.2_{2.3}$ &  & 86.9 & 53.6 & $66.3_{3.1}$ & 67.9 & 41.9 & $51.8_{3.4}$ \\

 & \sysname & 3 & 76.1 & 63.8 & $69.4_{2.6}$ & 66.5 & 55.7 & $\mathbf{60.6}_{3.1}$          &  & 87.3 & 65.3 & $74.7_{2.8}$ & 84.9 & 63.5 & $\mathbf{72.7}_{2.9}$          \\

\thickmid

\multirow{4}{*}{\textbf{Open}} &

Oracle rationale & 4 & 100.0 & 56.5 & $72.2_{3.3}$ & 87.6 & 49.5 & $63.2_{3.7}$ &  & 88.9 & 54.1 & $67.2_{3.2}$ & 88.9 & 54.1 & $67.2_{3.2}$ \\

\arrayrulecolor{black!30}\cmidrule(lr){2-16}
 & Zero-shot & 5 & 28.7 & 37.6 & $32.5_{2.3}$ & 23.7 & 31.1 & $26.9_{2.3}$ &  & 56.0 & 42.3 & $48.2_{3.3}$ & 42.3 & 32.0 & $36.4_{3.3}$ \\

 & \sysname & 6 & 45.0 & 47.3 & $46.1_{3.0}$ & 38.6 & 40.5 & $\mathbf{39.5}_{3.0}$          &  & 47.5 & 47.3 & $47.4_{3.1}$ & 46.6 & 46.4 & $\mathbf{46.5}_{3.1}$          \\

\arrayrulecolor{black}\bottomrule

  \end{tabularx}

  \caption{Test set performance on \ours, according to the metrics from \S\ref{sec:evaluation}. For the ``Oracle abstract'' rows, the system is provided with gold evidence abstracts. ``Oracle rationale'' rows indicate that the gold rationales are provided as input. ``Zero-shot'' indicates zero-shot performance of a verification system trained on \fever. Additionally, standard deviations are reported as subscripts for all F1 scores. See Appendix \ref{sec:uncertainty} for standard deviations on all reported metrics.}
    \label{tbl:main_results}

\end{table*}

\vspace{.1cm} \noindent{\bf Model Inputs}
We examine the performance of ``claim-only'' and ``abstract-only'' models trained on \ours, using \robertalarge as the sentence encoder. The claim-only model makes label predictions based on the claim text alone, without access to evidence abstracts. The abstract-only model selects rationale sentences and makes label predictions without access to the claim.

\vspace{.1cm} \noindent{\bf Results} The results are shown in Table \ref{tbl:components}. For \componentthree, the best performance is achieved by training first on the large \fever dataset and then fine-tuning on the smaller in-domain \ours training set. To understand the benefits of \fever pretraining, we examined the claim / evidence pairs where the \fever + \ours - trained model made correct predictions but the \ours - trained model did not. In  36 / 44 of these cases, the \ours - trained model predicts \notenough. Thus pretraining on \fever appears to improve the model's ability to recognize textual entailment relationships between evidence and claim -- particularly relationships indicated by non-domain-specific cues like ``is associated with'' or ``has an important role in''.

For \componenttwo, training on \ours alone produces the best results. We examined the rationales that the \ours - trained model identified but the \fever - trained model missed, and found that they generally contain science-specific vocabulary. Thus, training on additional out-of-domain data provides little benefit.

\robertalarge exhibits the strongest performance on label prediction, while \scibert has a slight edge on rationale selection. The ``claim-only'' model exhibits very poor performance, which provides some reassurance that the claim negation procedure described in \S\ref{sec:claim_writing} does not introduce obvious statistical artifacts. Similarly, the poor performance of the ``abstract-only'' model indicates that the model needs access to the claim being verified in order to identify relevant evidence.

\subsection{Full task} \label{sec:main_results}
\noindent {\bf Experimental setup}
Based on the results from \S\ref{sec:pipeline_components}, we use the \componenttwo module trained on \ours only, and the \componentthree module trained on \fever + \ours for our final end-to-end system \sysname. Although \scibert performs slightly better on rationale selection, using \robertalarge for both \componenttwo and \componentthree gave the best full-pipeline performance on the dev set, so we use \robertalarge for both components. For the \componentone module, the best dev set full-pipeline performance was achieved by retrieving the top $k=3$ documents. 

\vspace{.1cm}\noindent{\bf Model comparisons}
We report performance of three model variants. For the ``Oracle rationale'' setting, the \componenttwo module is replaced by an oracle which outputs gold rationales for correctly retrieved documents, and no rationales for incorrect retrievals. The ``Zero-shot'' setting reports the zero-shot generalization performance of a model trained on \fever (the results on \snopes were slightly worse). \sysname reports the performance of our best system.

\vspace{.1cm}\noindent{\bf Results}
The results are shown in Table \ref{tbl:main_results}. In the oracle abstract setting, the abstract-level F1 scores are roughly comparable to label classification accuracies, 
and the \abstractLabelRationale score in Row 3 implies an end-to-end classification accuracy of roughly 70\%, given gold abstracts.

\begin{table*}[t]
  \setlength{\tabcolsep}{2pt}
  \footnotesize
  \centering
  \begin{tabular}{L{0.15\linewidth} L{0.12\linewidth} L{0.7\linewidth}}
    \toprule
    \textbf{Reasoning type} &  \multicolumn{2}{l}{\textbf{Example}} \\
    \midrule

    \multirow{4}{*}{\shortstack[l]{\textbf{Science} \\ \textbf{background}}} & \textbf{Claim:} & Rapamycin slows aging in \hl{fruit flies}. \\
                            & \textbf{Evidence:} & \emph{\dots feeding rapamycin to adult \hl{Drosophila} produces life span extension \dots} \\
                            & \textbf{Gold Verdict}: & \supportsCovid{\textbf{\supports}} \\
                            & \textbf{Reasoning:} & \hl{Drosophila} is a type of \hl{fruit fly}.  \\

    \midrule

    \multirow{4}{*}{\textbf{Directionality}} & \textbf{Claim:} & Inhibiting glucose-6-phospate dehydrogenase \hl{impairs} lipogenesis \\
                            & \textbf{Evidence:} & \emph{\dots suppression of 6PGD \hl{increased} lipogenesis} \\
                            & \textbf{Gold Verdict:} & \refutesCovid{\textbf{\refutes}} \\
                            & \textbf{Reasoning:} & A \hl{decrease} (not \hl{increase}) in lipogenesis would indicate lipogenesis \hl{impairment}.  \\

    \midrule


    \multirow{5}{*}{\shortstack[l]{\textbf{Numerical} \\ \textbf{reasoning}}} & \textbf{Claim:}&  Bariatric surgery \hl{improves} resolution of diabetes. \\
                            & \textbf{Evidence:} & \emph{Strong associations were found between bariatric surgery and the resolution of T2DM,} \\
    & & \emph{\hl{with a HR of 9.29 (95\% CI 6.84-12.62)}...} {} \\
                            & \textbf{Gold Verdict:} & \supportsCovid{\textbf{\supports}} \\
                            & \textbf{Reasoning:} & A \hl{HR (hazard ratio)} that is greater than 1 with 95\% confidence indicates \hl{improvement}. \\

    \midrule

    \multirow{5}{*}{\shortstack[l]{\textbf{Cause and} \\ \textbf{effect}}} & \textbf{Claim:} & Major vault protein (MVP) functions to \hl{decrease} tumor aggression. \\
                            & \textbf{Evidence:} & \emph{\hl{Knockout} of MVP leads to miR-193a accumulation...\hl{inhibiting} tumor progression}  \\
                            & \textbf{Gold Verdict:} & \refutesCovid{\textbf{\refutes}} \\
                            & \textbf{Reasoning:} & \hl{Knocking out} (removing) MVP \hl{inhibits} tumor progression $\rightarrow$ MVP \hl{increases} tumor \\
    & & aggression. \\

    \midrule

    \multirow{5}{*}{\shortstack[l]{\textbf{Coreference}}} & \textbf{Claim:} & \hl{Low saturated fat diets} have adverse effects on the development of infants \\
                            & \textbf{Evidence:} & \emph{Neurological development of children in the \hl{intervention group} was at least as good as ...} \\
    & & \emph{the control group}  \\
                            & \textbf{Gold Verdict:} & \refutesCovid{\textbf{\refutes}} \\
                            & \textbf{Reasoning:} & The \hl{intervention group} in this study was placed on a \hl{low saturated fat diet}. \\

    \bottomrule
  \end{tabular}
  \caption{Reasoning types required to verify \ours claims which are classified incorrectly by our modeling baseline. Words crucial for correct verification are \hl{highlighted}.}
  \label{tbl:error_analysis}

\end{table*}

Access to in-domain data during training clearly improves performance. Despite the small size of \ours, training on these data leads to relative improvements of 47\% on open \sentenceSelectionLabel, and 28\% on open \abstractLabelRationale over \fever alone (Row 6 vs. Row 5).
The three pipeline components make similar contributions to the overall model error. Replacing \componenttwo with an oracle leads to a roughly 20-point rise in \sentenceSelectionLabel F1 (Row 6 vs. Row 4). Replacing \componentone with an oracle as well leads to a gain of roughly 20 more points (Row 4 vs. Row 1).

Nearly all correctly-labeled abstracts are supported by at least one rationale. There is only a two-point difference in F1 between \abstractLabelOnly and \abstractLabelRationale in the oracle setting (Row 3), and a one-point difference in the open setting (Row 6). The differences between \sentenceSelectionOnly and \sentenceSelectionLabel are larger, caused by examples where the model finds the evidence but fails to predict its relationship to the claim. We examine these in \S\ref{sec:error_analysis}.

We evaluate the statistical robustness of our results by generating 10,000 bootstrap-resampled versions of the test set \cite{Dror2018TheHG} and computing the standard deviation of all performance metrics. Table \ref{tbl:main_results} shows the standard deviations in F1 score. Uncertainties on all metrics for both the dev and test set can be found in Appendix \ref{sec:uncertainty}. The results indicate that the observed differences in model performance are statistically robust and cannot be attributed to random variation in the dataset.


\subsection{Verifying claims about COVID-19} \label{sec:covid}
We conduct exploratory experiments using our system to verify claims concerning COVID-19. We tasked a medical student to write 36 COVID-related claims. For each claim $c$, we used \sysname to predict evidence abstracts $\Ehat(c)$. The annotator examined each $(c, \Ehat(c))$ pair. A pair was labeled \emph{plausible} if $\Ehat(c)$ was nonempty, and at least half of the evidence abstracts in $\Ehat(c)$ were judged to have reasonable rationales and labels. For 23 / 36 claims, the response of \sysname was deemed plausible by our annotator, demonstrating that \sysname is able to successfully retrieve and classify evidence in many cases. Two examples are shown in Table \ref{tbl:covid_example}. In both cases, our system identifies both supporting \emph{and} refuting evidence.


\subsection{Error analysis} \label{sec:error_analysis}

To better understand the errors made by \sysname, we conduct a manual analysis of test set predictions where an evidence abstract was correctly retrieved, but where the model failed to identify any relevant rationales or predicted an incorrect label. We identify five modeling capabilities required to correct these mistakes (Table \ref{tbl:error_analysis} provides examples):

\begin{description}[leftmargin=*,leftmargin=0pt]
  \item \textbf{Science background} includes knowledge of domain-specific lexical relationships.
  \item \textbf{Directionality} requires understanding increases or decreases in scientific quantities.
  \item \textbf{Numerical reasoning} involves interpreting numerical or statistical findings.
  \item \textbf{Cause and effect} requires reasoning about counterfactuals.
  \item \textbf{Coreference} involves drawing conclusions using context stated outside of a rationale sentence.
\end{description}

\section{Related work}

\label{sec:related_work}

\noindent {\bf Fact checking and rationalized NLP models}
Fact-checking datasets include PolitiFact \cite{Vlachos2014FactCT}, Emergent \cite{ferreira-vlachos-2016-emergent}, LIAR \cite{Wang2017LiarLP}, SemEval 2017 Task 8 RumorEval \cite{derczynski-etal-2017-semeval}, Snopes \cite{Popat2017WhereTT}, CLEF-2018 CheckThat! \cite{nakov2018overview}, Verify \cite{baly-etal-2018-integrating}, Perspectrum \cite{Chen2019SeeingTF}, \fever \cite{Thorne2018FEVERAL}, and \snopes \cite{Hanselowski2019ARA}. \citet{Hanselowski2019ARA} provides a thorough review.  To our knowledge, there are no existing data sets for scientific claim verification. We refer to our task as ``claim verification'' rather than ``fact-checking'' to emphasize that our focus is to help researchers make sense of scientific findings, not to counter disinformation.

Fact-checking is one of a number of tasks where a model is required to justify a prediction via \emph{rationales} from the source document. The ERASER dataset \cite{DeYoung2019ERASERAB} provides a suite of benchmark datasets (including \ours) for evaluating rationalized NLP models.


\vspace{.1cm} \noindent {\bf Related scientific NLP tasks}
\label{sec:related_scinlp}
The \emph{citation contextualization} task \cite{Cohan2015MatchingCT,Jaidka2017TheCS} is to identify spans in a cited document that are relevant to a particular citation in a citing document. Unlike \ours, these citations are not re-written into atomic claims and are therefore more difficult to verify. Expert annotators achieved very low (21.7\%) inter-annotator agreement on the BioMedSumm dataset \cite{BiomedSumm}, which contains 314 citations referencing 20 papers.

\emph{Biomedical question answering} datasets include BioASQ \cite{Tsatsaronis2015AnOO} and PubMedQA \cite{Jin2019PubMedQAAD}, which contain 855 and 1,000 ``\emph{yes / no}'' questions respectively \cite{Gu2020DomainSpecificLM}. Claim verification and question answering are both-knowledge intensive tasks which require an understanding of the relationship between an input query and relevant supporting text.

\emph{Automated evidence synthesis} \cite{Marshall2019TowardSR, Beller2018MakingPW, Tsafnat2014SystematicRA, Marshall2017AutomatingBE} seeks to automate the process of creating \emph{systematic reviews} of the medical literature\footnote{\url{https://www.cochranelibrary.com/about/about-cochrane-reviews}} -- for instance, by extracting PICO snippets \cite{Nye2018ACW} and inferring the outcomes of clinical trials \cite{Lehman2019InferringWM, DeYoung2020EvidenceI2}. We hope that systems for claim verification will serve as components in future evidence synthesis frameworks.
\section{Conclusion and future work}

Claim verification allows us to trace the sources and measure the veracity of scientific claims. These abilities have emerged as particularly important in the context of the current pandemic, and the broader reproducibility crisis in science. In this article, we formalize the task of scientific claim verification, and release a dataset (\ours) and models (\sysname) to support work on this task. Our results indicate that it is possible to train models for scientific fact-checking and deploy them with reasonable efficacy on real-world claims related to COVID-19.

Scientific claim verification presents a number of promising avenues for research on models capable of incorporating background information, reasoning about scientific processes, and assessing the strength and provenance of various evidence sources. This last challenge will be especially crucial for future work that seeks to verify scientific claims against sources other than the research literature -- for instance, social media and the news. We hope that the resources presented in this paper encourage future research on these important challenges, and help facilitate progress toward the broader goal of scientific document understanding.

\section*{Acknowledgments}

This research was supported by the ONR MURI N00014-18-1-2670, ONR N00014-18-1-2826, DARPA N66001-19-2-4031, NSF (IIS 1616112), Allen Distinguished Investigator Award, and the Sloan fellowship. We thank the Semantic Scholar team at AI2, UW-NLP, and H2lab at  UW for helpful comments and feedback.

\clearpage

\bibliography{emnlp2020}
\bibliographystyle{acl_natbib}
\clearpage

\appendix

\section{Model implementation details} \label{sec:model_details}

All models are implemented using the Huggingface Transformers package \cite{Wolf2019HuggingFacesTS}.


\subsection{Parameters for the final \sysname system} \label{sec:parameters}

For the \componentone module, \sysname retrieves the top $k=3$ documents ranked by TF-IDF similarity using unigram + bigram features. These parameters are tuned on the \ours development set.

When making predictions using the \componenttwo module described in \S\ref{sec:rationale_selection}, we find that the usual decision rule of predicting $\hat{z}_i = 1$ when $\tilde{z}_i \geq 0.5$ works well for models trained on \ours. However, for models trained on \fever and \snopes, we achieve better performance by tuning the classification threshold $t$, such that $\hat{z}_i = 1$  when $\tilde{z}_i \geq t$, on the \ours dev set. The best threshold was $t=0.025$ when training on \fever, and $t=0.75$ when training on \snopes.


\subsection{Training the \componenttwo module}

We experiment with various learning rates when training \scibert, \bioroberta, \robertabase, and \robertalarge. Below we describe the setting for training \robertalarge.

For models trained on \ours, we use an initial learning rate of 1e-5 on the transformer base and 1e-3 on the linear layer. For \fever + \ours, the learning rate is set to 1e-5 for the entire model for pre-training on \fever and fine-tuning on \ours. We use a batch size of 256 through gradient accumulation and apply cosine learning rate decay over 20 epochs to find the best performing model on the dev set.

For models trained on \fever, we set the learning rate to 5e-6 for the transformer base and 5e-5 for the linear layer. For models trained on \snopes, we set the learning rate 1e-5 for the transformer base and 1e-4 for the linear layer. We find that these learning rates help the models converge. We only train the model for 3 epochs on \fever and 5 epochs on \snopes because they are larger datasets and the models converged within early epochs.


\subsection{Training the \componentthree module}

We adopt similar settings as we used for the \componenttwo module and only change the learning rate to 1e-5 for the transformer base and 1e-4 for the linear layer for models trained on \ours, \fever, and \snopes. When training on claim / cited abstract pairs labeled \notenough, we use the $k$ sentences in the abstract with greatest similarity to the claim as rationales (\S\ref{sec:baselines}). $k$ is sampled from $\{0, 1\}$ with uniform probability.


\subsection{Additional training details}

All models are trained using a single Nvidia P100 GPU on Google Colabortoary Pro platform.\footnote{\url{https://colab.research.google.com/}} For the \componenttwo module, it takes about 150 minutes to train on \ours for 20 epochs. 120 minutes on \snopes for 5 epochs, and 700 minutes on \fever for 3 epochs. For the \componentthree module, it takes about 130 minutes to train on \ours for 20 epochs, 160 minutes on \snopes for 5 epochs, and 640 minutes on \fever for 3 epochs.


\subsection{Hyperparameter search}

The learning rate, batch size, and number of epochs are the most important hyperparameters. We perform manual tuning and select the hyperparameters that produce the highest F1 on the development set. For the learning rate, we experiment with 1e-3, 1e-4, 5e-5, 1e-5, and 5e-6. For batch size, we experiment with 64 and 256. The number of epochs are cutoff after the model converges.


\section{Statistical analysis} \label{sec:uncertainty}

We assess the uncertainty in the results reported in the main results (Table \ref{tbl:main_results}) using a simple bootstrap approach \cite{Dror2018TheHG, BergKirkpatrick2012AnEI, Efron1993AnIT}. Given our test set with $n_{\textrm{test}} = 300$ claims, we generate $n_{\textrm{boot}} = 10,000$ bootstrap-resampled test sets by resampling (uniformly, with replacement) $n_{\textrm{test}}$ claims from the test set. For each resampled test set, we compute the metrics in Table \ref{tbl:main_results}. Table \ref{tbl:results_test_bootstrap} reports the mean and standard deviation of these metrics, computed over the bootstrap samples. Table \ref{tbl:results_dev_bootstrap} reports dev set metrics. Our conclusion that training on \ours improves performance is robust to the uncertainties presented in these tables.

\clearpage

\begin{table*}[t]
  \footnotesize
  \setlength{\tabcolsep}{0.5em}

  \begin{subtable}[h]{\linewidth}
    \centering
    \begin{tabularx}{0.8\linewidth}{*{3}{l} *{6}{c}}

      \toprule

      & & & \multicolumn{6}{c}{\textbf{Sentence-level}} \\
      & & & \multicolumn{3}{c}{\textbf{Selection-Only}} & \multicolumn{3}{c}{\textbf{Selection+Label}} \\

      \cmidrule(lr){4-6} \cmidrule(lr){7-9}

      Retrieval & Model & Row & P & R & F1 & P & R & F1 \\
      \midrule
      \multirow{4}{*}{\shortstack[c]{\textbf{Oracle} \\ \textbf{abstract}}} &
      Oracle rationale & 1 & $100.0_{0.0}$ & $80.5_{3.3}$ & $89.2_{2.1}$ & $89.6_{2.7}$ & $72.2_{3.7}$ & $79.9_{3.0}$ \\

      \arrayrulecolor{black!30}\cmidrule(lr){2-9}
      & Zero-shot & 2 & $42.6_{2.2}$ & $45.2_{3.2}$ & $43.8_{2.0}$ & $36.2_{2.5}$ & $38.4_{3.0}$ & $37.2_{2.3}$ \\
      & \sysname & 3 & $76.2_{2.9}$ & $63.9_{3.6}$ & $69.4_{2.6}$ & $66.5_{3.4}$ & $55.7_{3.7}$ & $60.6_{3.1}$ \\

      \thickmid

      \multirow{4}{*}{\textbf{Open}} &
      Oracle rationale               & 4 & $100.0_{0.0}$ & $56.6_{4.0}$ & $72.2_{3.3}$ & $87.6_{3.5}$ & $49.5_{3.9}$ & $63.2_{3.7}$ \\
      \arrayrulecolor{black!30}\cmidrule(lr){2-9}
      & Zero-shot & 5 & $28.7_{2.3}$ & $37.6_{3.4}$ & $32.5_{2.3}$ & $23.8_{2.3}$ & $31.1_{3.1}$ & $26.9_{2.3}$ \\
      & \sysname & 6 & $45.0_{3.0}$ & $47.4_{3.8}$ & $46.1_{3.0}$ & $38.5_{3.0}$ & $40.6_{3.6}$ & $39.5_{3.0}$ \\

    \end{tabularx}
    \subcaption{Sentence-level results.}
  \end{subtable}

  \begin{subtable}[h]{\linewidth}
    \centering
    \begin{tabularx}{0.8\linewidth}{*{3}{l} *{6}{c}}

      \toprule

      &  &  & \multicolumn{6}{c}{\textbf{Abstract-level}} \\
      &  &  & \multicolumn{3}{c}{\textbf{Label-Only}} & \multicolumn{3}{c}{\textbf{Label+Rationale}} \\

      \cmidrule(lr){4-6} \cmidrule(lr){7-9}

      Retrieval & Model & Row & P & R & F1 & P & R & F1 \\

      \midrule

      \multirow{4}{*}{\shortstack[c]{\textbf{Oracle} \\ \textbf{abstract}}} &
      Oracle rationale & 1 & $90.1_{2.2}$ & $77.5_{2.8}$ & $83.3_{2.4}$ & $90.1_{2.2}$ & $77.5_{2.8}$ & $83.3_{2.4}$ \\

      \arrayrulecolor{black!30}\cmidrule(lr){2-9}

      & Zero-shot & 2 & $86.9_{2.9}$ & $53.6_{3.4}$ & $66.3_{3.1}$ & $67.9_{3.9}$ & $41.9_{3.2}$ & $51.8_{3.4}$ \\
      & \sysname & 3 & $87.3_{2.6}$ & $65.3_{3.2}$ & $74.7_{2.8}$ & $84.9_{2.8}$ & $63.5_{3.2}$ & $72.6_{2.9}$ \\

      \thickmid

      \multirow{4}{*}{\textbf{Open}} &

      Oracle rationale               & 4 & $88.9_{2.7}$ & $54.1_{3.5}$ & $67.2_{3.2}$ & $88.9_{2.7}$ & $54.1_{3.5}$ & $67.2_{3.2}$ \\

      \arrayrulecolor{black!30}\cmidrule(lr){2-9}
      & Zero-shot & 5 & $56.0_{3.9}$ & $42.3_{3.4}$ & $48.2_{3.3}$ & $42.3_{4.0}$ & $32.0_{3.2}$ & $36.4_{3.3}$ \\
      & \sysname & 6  &  $47.5_{3.3}$ & $47.3_{3.5}$ & $47.4_{3.1}$ & $46.6_{3.3}$ & $46.4_{3.5}$ & $46.4_{3.1}$ \\

      \arrayrulecolor{black}\bottomrule

    \end{tabularx}
    \subcaption{Abstract-level results}
  \end{subtable}

  \caption{Test set results as in Table \ref{tbl:main_results}, reporting mean and standard deviation over 10,000 bootstrap samples. Standard deviations are reported as subscripts. Some means reported here are slightly different from Table \ref{tbl:main_results} due to sampling variability.}

  \label{tbl:results_test_bootstrap}

\end{table*}
\begin{table*}[h!]
  \footnotesize
  \setlength{\tabcolsep}{0.5em}

  \begin{subtable}[h]{\linewidth}
    \centering
    \begin{tabularx}{0.8\linewidth}{*{3}{l} *{6}{c}}

      \toprule

      & & & \multicolumn{6}{c}{\textbf{Sentence-level}} \\
      & & & \multicolumn{3}{c}{\textbf{Selection-Only}} & \multicolumn{3}{c}{\textbf{Selection+Label}} \\

      \cmidrule(lr){4-6} \cmidrule(lr){7-9}

      Retrieval & Model & Row & P & R & F1 & P & R & F1 \\
      \midrule
      \multirow{4}{*}{\shortstack[c]{\textbf{Oracle} \\ \textbf{abstract}}} &
      Oracle rationale & 1 &   $100.0_{0.0}$ &       $81.9_{3.2}$ &       $90.0_{1.9}$ &   $91.4_{2.5}$ &   $74.9_{3.6}$ &   $82.3_{2.9}$ \\

      \arrayrulecolor{black!30}\cmidrule(lr){2-9}
      & Zero-shot & 2 &  $40.7_{2.1}$ &       $48.1_{3.4}$ &       $44.0_{2.1}$ &   $36.1_{2.5}$ &   $42.6_{3.4}$ &   $39.0_{2.5}$ \\
      & \sysname & 3 &  $79.4_{2.7}$ &       $59.0_{3.6}$ &       $67.7_{2.8}$ &   $71.4_{3.5}$ &   $53.0_{3.6}$ &   $60.8_{3.3}$ \\

      \thickmid

      \multirow{4}{*}{\textbf{Open}} &
      Oracle rationale               & 4 &   $100.0_{0.0}$ &       $58.4_{4.3}$ &       $73.7_{3.4}$ &   $90.2_{3.3}$ &   $52.7_{4.3}$ &   $66.4_{3.9}$ \\
      \arrayrulecolor{black!30}\cmidrule(lr){2-9}
      & Zero-shot & 5 &   $28.6_{2.0}$ &       $38.5_{3.6}$ &       $32.8_{2.3}$ &   $24.8_{2.2}$ &   $33.4_{3.4}$ &   $28.4_{2.4}$ \\
      & \sysname & 6 &       $52.5_{3.5}$ &       $43.8_{3.7}$ &       $47.7_{3.2}$ &   $46.9_{3.7}$ &   $39.2_{3.6}$ &   $42.6_{3.2}$ \\

    \end{tabularx}
    \subcaption{Sentence-level results.}
  \end{subtable}

  \begin{subtable}[h]{\linewidth}
    \centering
    \begin{tabularx}{0.8\linewidth}{*{3}{l} *{6}{c}}

      \toprule

      &  &  & \multicolumn{6}{c}{\textbf{Abstract-level}} \\
      &  &  & \multicolumn{3}{c}{\textbf{Label-Only}} & \multicolumn{3}{c}{\textbf{Label+Rationale}} \\

      \cmidrule(lr){4-6} \cmidrule(lr){7-9}

      Retrieval & Model & Row & P & R & F1 & P & R & F1 \\

      \midrule

      \multirow{4}{*}{\shortstack[c]{\textbf{Oracle} \\ \textbf{abstract}}} &
      Oracle rationale & 1 &  $91.4_{2.2}$ &   $76.1_{3.0}$ &   $83.0_{2.5}$ &       $91.4_{2.2}$ &       $76.1_{3.0}$ &       $83.0_{2.5}$ \\

      \arrayrulecolor{black!30}\cmidrule(lr){2-9}

      & Zero-shot & 2 &    $88.9_{2.8}$ &   $58.3_{3.7}$ &   $70.4_{3.2}$ &       $69.2_{3.9}$ &       $45.4_{3.5}$ &       $54.8_{3.5}$ \\
      & \sysname & 3 &    $91.0_{2.3}$ &   $67.4_{3.3}$ &   $77.4_{2.7}$ &       $85.2_{2.9}$ &       $63.2_{3.5}$ &       $72.5_{3.1}$ \\

      \thickmid

      \multirow{4}{*}{\textbf{Open}} &

      Oracle rationale               & 4 &    $91.0_{2.6}$ &   $53.1_{3.8}$ &   $67.0_{3.4}$ &       $91.0_{2.6}$ &       $53.1_{3.8}$ &       $67.0_{3.4}$ \\

      \arrayrulecolor{black!30}\cmidrule(lr){2-9}
      & Zero-shot & 5 &   $52.7_{3.7}$ &   $41.6_{3.7}$ &   $46.5_{3.4}$ &       $43.6_{3.7}$ &       $34.4_{3.5}$ &       $38.4_{3.3}$ \\
      & \sysname & 6  &   $55.4_{3.7}$ &   $47.5_{3.6}$ &   $51.0_{3.3}$ &       $52.6_{3.7}$ &       $45.1_{3.6}$ &       $48.5_{3.3}$ \\

      \arrayrulecolor{black}\bottomrule

    \end{tabularx}
    \subcaption{Abstract-level results}
  \end{subtable}

  \caption{Dev set results as in Table \ref{tbl:main_results}, reporting mean and standard deviation over 10,000 bootstrap samples.}

  \label{tbl:results_dev_bootstrap}

\end{table*}

\clearpage


\section{Dataset collection and corpus statistics} \label{sec:data_collection}

\subsection{Corpus} \label{sec:data_source_appendix}

\paragraph{Source journals} Table \ref{tbl:journal_counts} shows the number of cited abstracts from each of our selected journals. The ``Other'' category includes ``co-cited'' (\S\ref{sec:data_source}) abstracts that came from journals not among our pre-defined set.

\begin{table}[t]
  \centering

  \begin{tabular}{l r}
  \toprule
  Journal & Count \\
  \thickmid
  BMJ                            &     60 \\
Blood                          &      8 \\
Cancer Cell                    &      8 \\
Cell                           &     51 \\
Cell Metabolism                &     10 \\
Cell Stem Cell                 &     41 \\
Circulation                    &     12 \\
Immunity                       &     33 \\
JAMA                           &     79 \\
Molecular Cell                 &     27 \\
Molecular Systems Biology      &      5 \\
Nature                         &     29 \\
Nature Cell Biology            &     26 \\
Nature Communications          &     19 \\
Nature Genetics                &      8 \\
Nature Medicine                &     89 \\
Nature Methods                 &      1 \\
Nucleic Acids Research         &     10 \\
Plos Biology                   &     36 \\
Plos Medicine                  &     38 \\
Science                        &      7 \\
Science Translational Medicine &      2 \\
The Lancet                     &     22 \\
\thinmid
Other                          &    120 \\

\thickmid

Total                          &    741 \\
\bottomrule
\end{tabular}

  \caption{Number of cited documents by journal. Some co-cited articles (\S \ref{sec:data_source}) come from journals outside our curated set; these are indicated by ``Other''.}
  \label{tbl:journal_counts}
\end{table}

\paragraph{Distractor abstracts} In \S\ref{sec:data_source}, we mention how we increase the size of the corpus by adding \emph{distractor abstracts}. The reason why we do not use the entirety of a large research corpus like S2ORC as our fact-checking corpus is that doing so would introduce many \emph{false negative retrievals}: abstracts containing evidence relevant to a given claim, but not mentioned in the claim's source citance. This can occur either because the citance authors simply were not aware of these abstracts, or because the abstracts were published after the citance was written. These retrievals would be incorrectly marked wrong by our evaluation metrics.

Distractor abstracts as defined in \S\ref{sec:data_source} have two qualities that make them a good addition to the \ours corpus: (1) They are cited in the same articles as our evidence abstracts, meaning that they often discuss similar topics and increase the difficulty of abstract retrieval methods based on lexical similarity. (2) The authors of our citances were aware of the distractor abstracts, and chose not to mention them in the citances used to generate claims. This makes them unlikely to be a source of false negative retrievals.


\subsection{Annotation examples} \label{sec:annotation_examples}

\begin{figure}[t]
  \centering
  \includegraphics[width=\columnwidth]{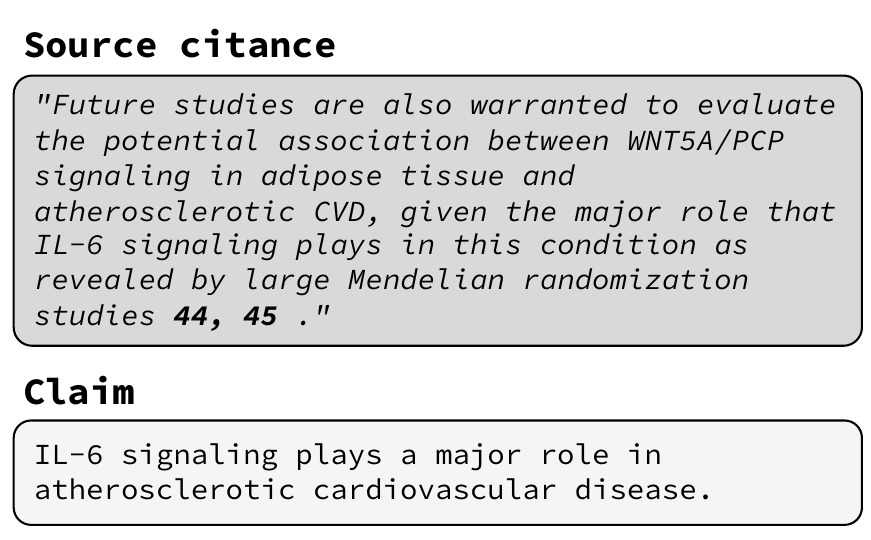}

  \caption{A claim written based on a citance. Material unrelated to the citation is removed. The acronym ``CVD'' is expanded to ``cardiovascular disease''.}
  \label{fig:claim}
\end{figure}

\paragraph{Converting citances to claims} Figure \ref{fig:claim} shows an example of a citance re-written as a claim. The citance discusses the relationship between ``atherosclerotic CVD'' and ``IL-6'', and cites two papers (\textbf{44} and \textbf{45}) as evidence. To convert to a claim, the acronym ``CVD'' is expanded to ``cardiovascular disease'', irrelevant information is removed, and the claim is written as an atomic factual statement.

\begin{figure}[t]
  \centering

  \includegraphics[width=\columnwidth]{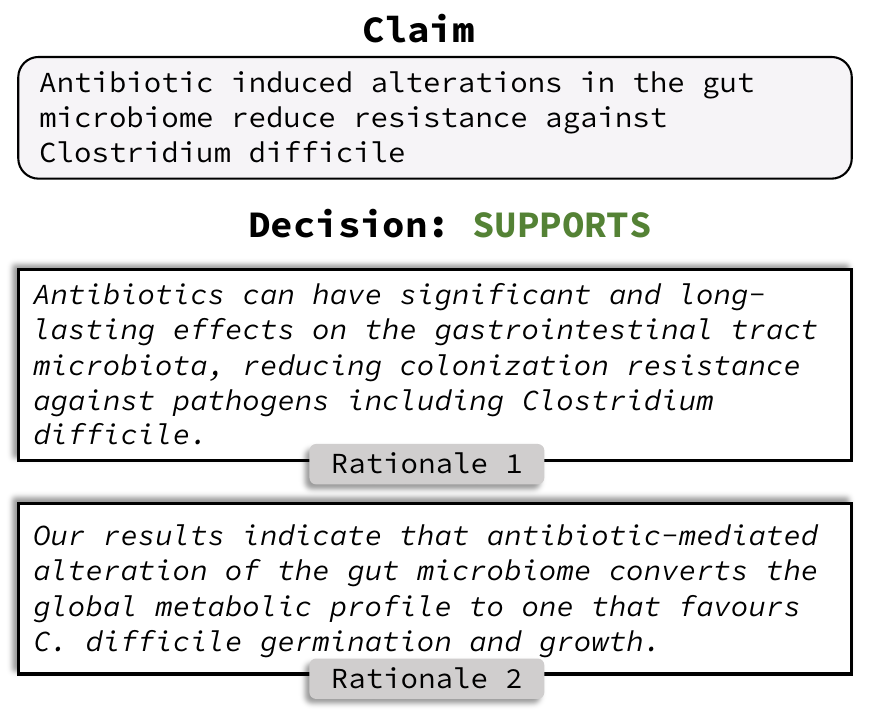}
  \caption{A claim supported by two rationales from the same abstract. The text of each rationale on its own provides sufficient evidence to verify the claim.}
  \label{fig:multiple_rationales}

\end{figure}

\paragraph{Multiple rationales} Figure \ref{fig:multiple_rationales} shows a claim supported by two rationales from the same abstract. The text of each rationale on its own is sufficient to entail the claim.

\subsection{Annotators and quality control} \label{sec:annotator_training}

\paragraph{Claim writing} Student claim writers attended an in-person training session where they were introduced to the task and received in-person feedback from the four experts. Following training, student annotators continued writing claims remotely. The expert annotators monitored claims for quality during the remote annotation process, and provided feedback when necessary; low-quality claims were returned to the annotators for re-writing. As a final check, all submitted claims were proofread (and edited if necessary) by an undergraduate whose claims were deemed especially high-quality by the expert annotators.

\paragraph{Claim negations}

As mentioned in \S\ref{sec:claim_writing}, an expert annotator wrote claim negations to introduce cases where an abstract \refutes a claim. The annotator skipped claims that could only be negated by adding obvious triggers like ``not''. The majority of claim negations involved a reversal of effect direction; for instance ``\emph{A high microerythrocyte count protects against severe anemia}'' can be negated as ``\emph{A high microerythrocyte count raises vulnerability to severe anemia}''.

\paragraph{Claim verification}

Annotations were performed remotely through a web interface. Annotators were required to pass a 10-question ``quiz'' before annotating their own claims. After passing the quiz, subsequent submissions were reviewed by an NLP expert until that expert deemed the annotator reliable. Approved annotators were then assigned to review each others' submissions. In general, graduate students were assigned to review annotations from undergraduates.


\section{Annotation interfaces and guidelines} \label{sec:annotation_interfaces}

\begin{figure*}[t]
  \centering
  \includegraphics[width=\textwidth]{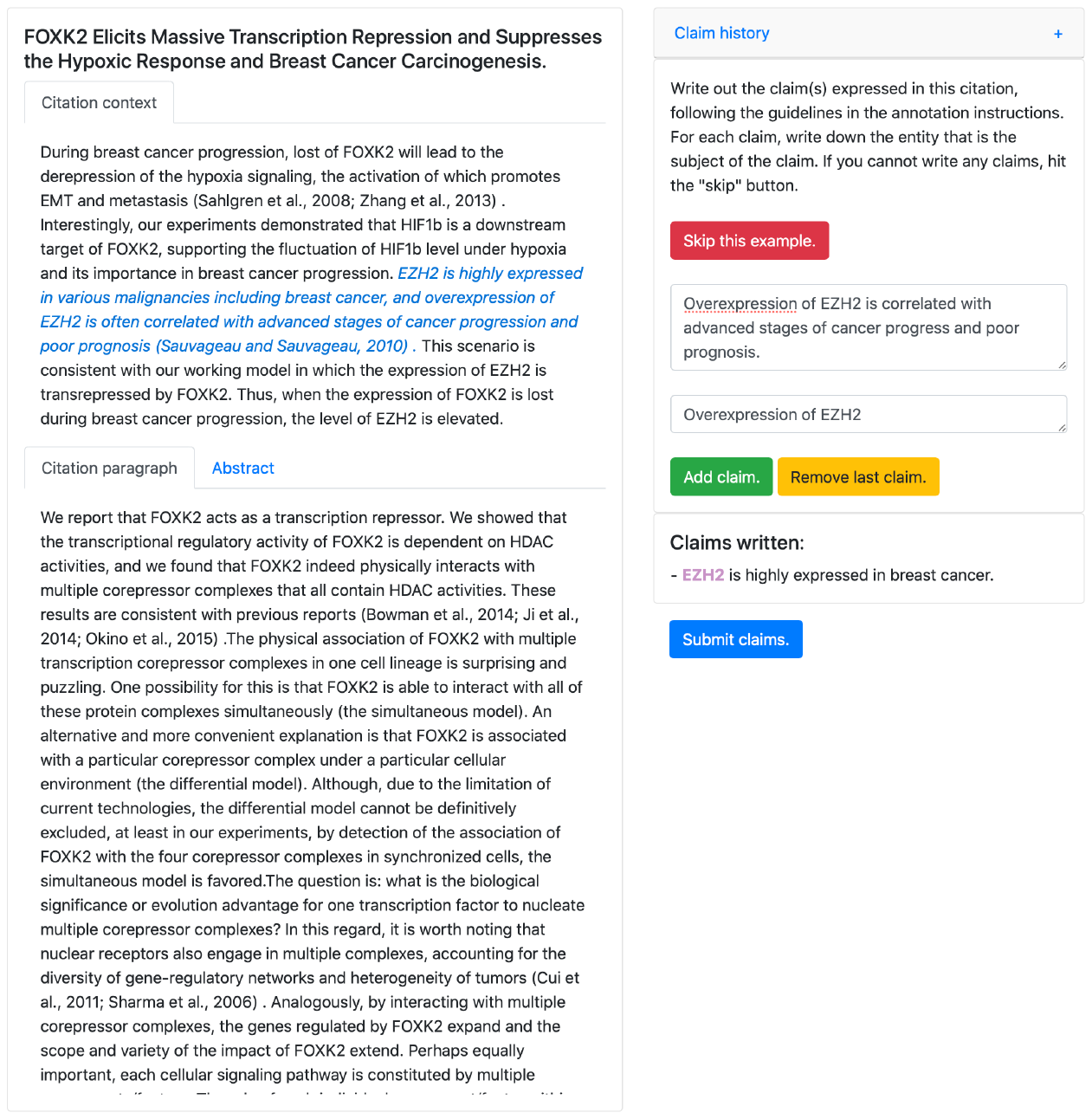}

  \caption{The claim-writing interface. The citation sentence is highlighted in blue on the top left. Additional context is provided on bottom left. The right side shows two claims that could be written based on this citation sentence.}
  \label{fig:claim_interface}
\end{figure*}

\begin{figure*}[h]
  \centering
  \includegraphics[width=\textwidth]{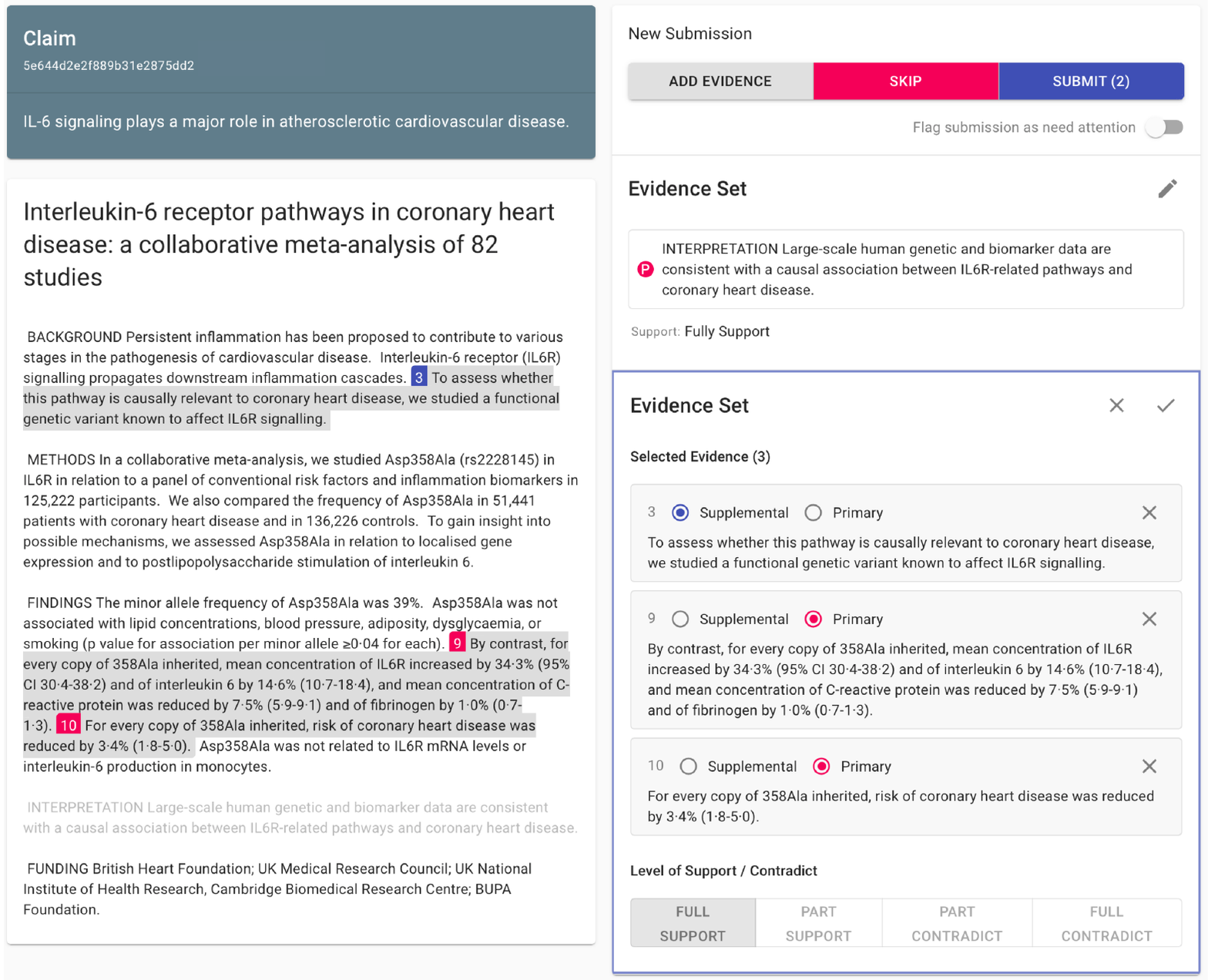}

  \caption{The evidence collection interface.}
  \label{fig:evidence_interface}
\end{figure*}

We show a screenshot of the claim writing interface in Figure \ref{fig:claim_interface}, and the claim verification interface in Figure \ref{fig:evidence_interface}. The complete annotation guide for claim verification is available at the following URL: \url{https://scifact.s3-us-west-2.amazonaws.com/doc/evidence-annotation-instructions.pdf}.

\end{document}